\documentclass[10pt,twocolumn,letterpaper]{article}

\usepackage{cvpr}              

\usepackage{graphicx}
\usepackage{amsmath}
\usepackage{amssymb}
\usepackage{booktabs}

\usepackage[pagebackref,breaklinks,colorlinks]{hyperref}
\usepackage{multirow}
\usepackage{float}

\usepackage[capitalize]{cleveref}
\crefname{section}{Sec.}{Secs.}
\Crefname{section}{Section}{Sections}
\Crefname{table}{Table}{Tables}
\crefname{table}{Tab.}{Tabs.}

\def\cvprPaperID{*****} 
\def\confName{CVPR}
\def\confYear{2022}

\begin{document}

\title{Technical Report for \confName~2022 LOVEU AQTC Challenge}

\author{Hyeonyu Kim$^*$\\
UNIST\\
{\tt\small khy0501@unist.ac.kr}
\and
Jongeun Kim$^*$\\
UNIST\\
{\tt\small joannekim0420@unist.ac.kr}
\and
Jeonghun Kang\\
UNIST\\
{\tt\small jhkang@unist.ac.kr}
\and
Sanguk Park\\
Pyler, Inc.\\
{\tt\small psycoder@pyler.tech}
\and
Dongchan Park\\
Pyler, Inc.\\
{\tt\small cto@pyler.tech}
\and
Taehwan Kim\\
UNIST\\
{\tt\small taehwankim@unist.ac.kr}
}

\maketitle
\def\thefootnote{*}\footnotetext{These authors contributed equally to this work}
\begin{abstract}
   This technical report presents the 2nd winning model for AQTC, a task newly introduced in CVPR 2022 LOng-form VidEo Understanding (LOVEU) challenges. This challenge faces difficulties with multi-step answers, multi-modal, and diverse and changing button representations in video. We address this problem by proposing a new context ground module attention mechanism for more effective feature mapping. In addition, we also perform the analysis over the number of buttons and ablation study of different step networks and video features. As a result, we achieved the overall 2nd place in LOVEU competition track 3, specifically the 1st place in two out of four evaluation metrics.
   Our code is available at \url{https://github.com/jaykim9870/CVPR-22_LOVEU_unipyler}.
\end{abstract}

\section{Introduction}

\begin{figure}[t!]
  \centering \includegraphics[width=\linewidth]{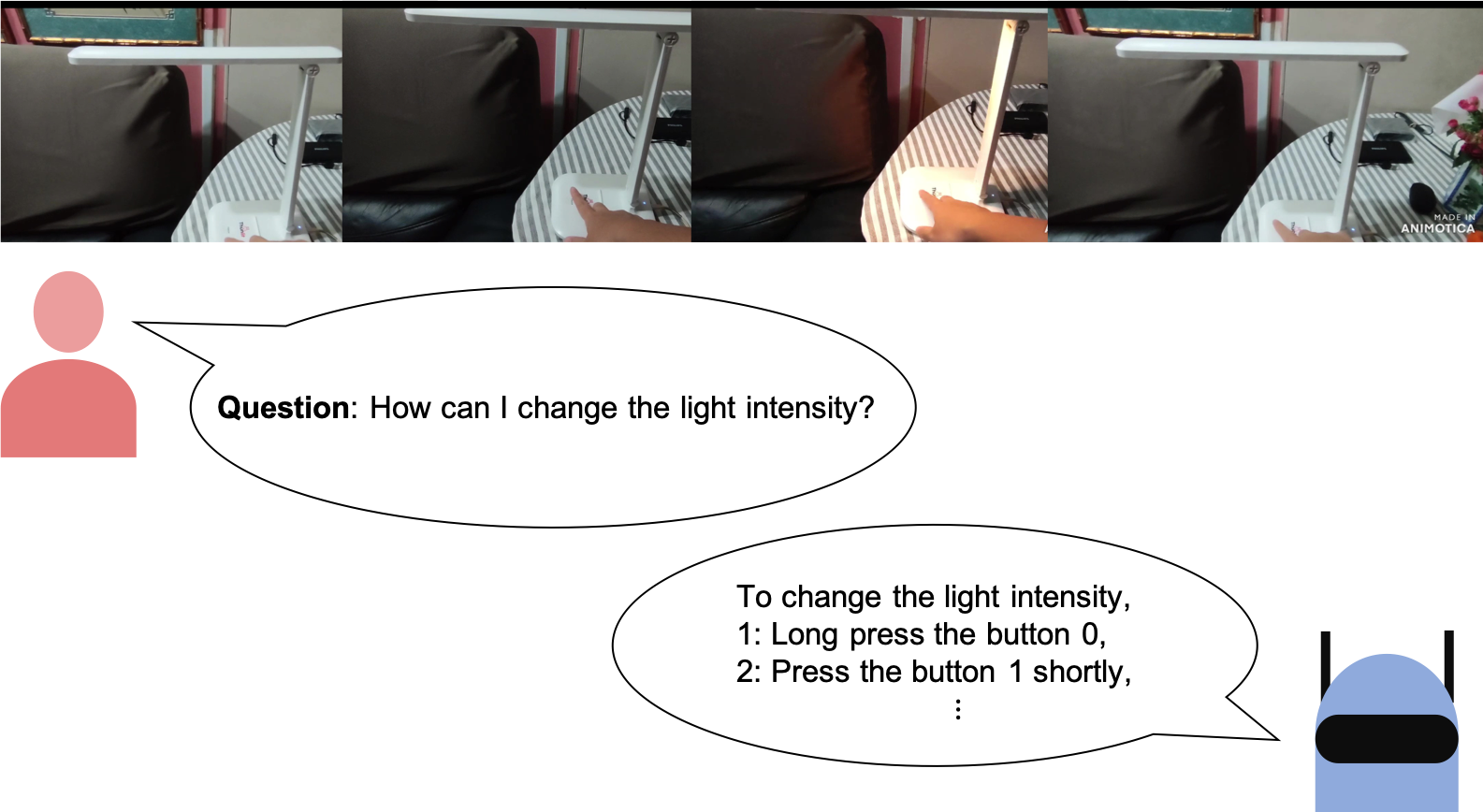}
  \caption{An example of an AQTC task in AssistQ dataset.}
  \label{fig:data_ex} 
\end{figure}

Affordance-centric Question-driven Task Completion (AQTC) is a new challenge introduced in CVPR 2022 LOVEU workshop. The task aims to help users operate machines in the real-world from an affordance-centric perspective by answering users' questions such as, "How to turn on the light stand?". To guide the users step-by-step to achieve their goals, intelligent assistants should learn from the instructional videos and scripts from the users' point of view \cite{wong2022assistq}. To help building such AI agents, AssistQ dataset was created consisting of instruction videos with scripts and questions and their answers. An example of dataset on how to operate a lightstand is illustrated in Figure \ref{fig:data_ex}. 
However, this task is more demanding than other question answering tasks like Visual Question Answering \cite{antol2015vqa}, Embodied Question Answering \cite{das2018embodied, yu2019multi}, and Visual Dialog \cite{alamri2019audio,das2017visual} for three reasons. First, AQTC requires multiple-step answers in each question with multiple-choice in each answer. Second, the buttons may look visually different during step-by-step actions by egocentric movement. Lastly, AQTC deals with multiple modalities of text and image for videos and scripts.
\par To solve AQTC, we propose a new context ground module attention mechanism for more effective feature mapping. Additionally, we also perform ablation study to investigate the effect of different step networks and video sampling rates.



\section{Our Model}
\begin{figure*}[h]
  \centering \includegraphics[width=\textwidth]{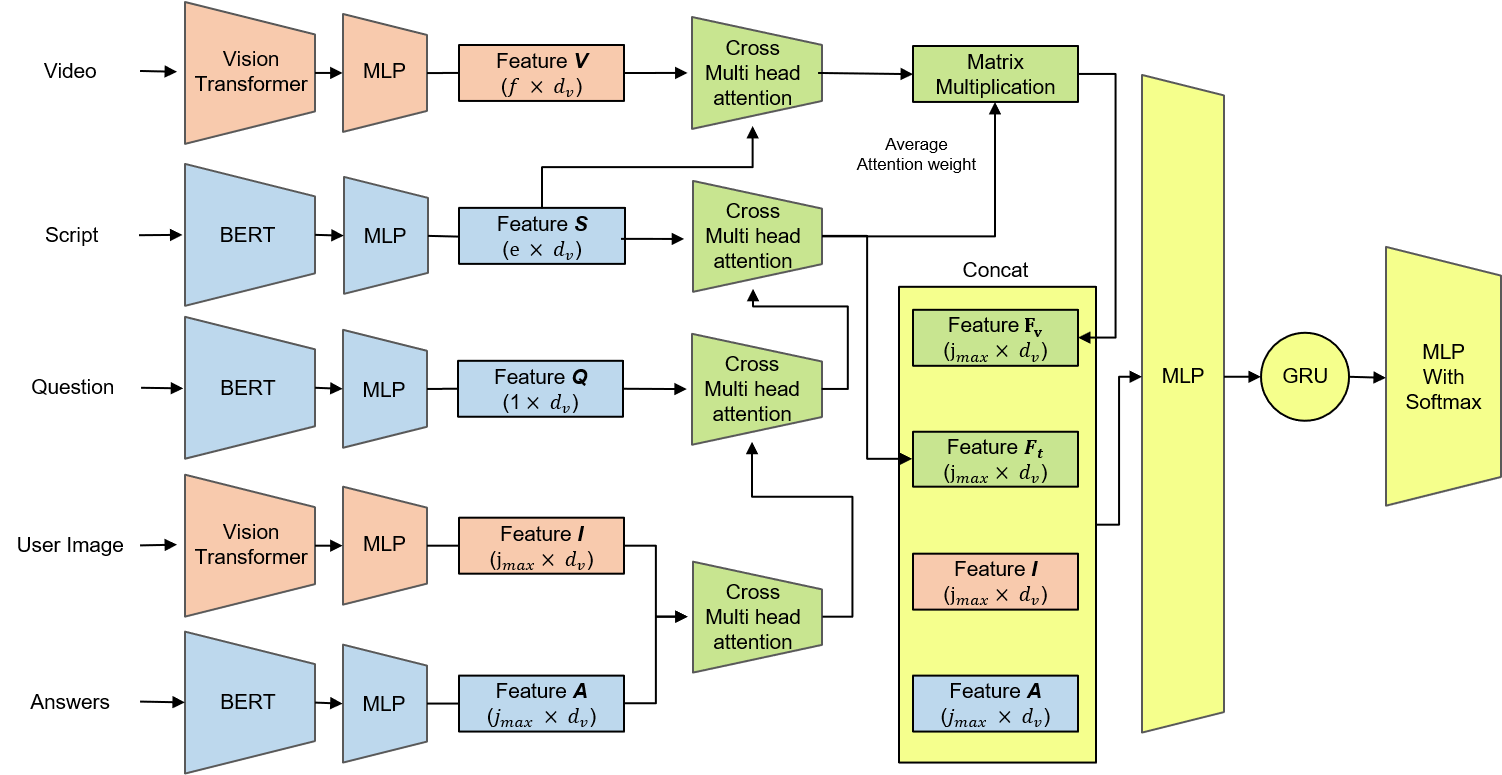}
  \caption{Model architecture. Each feature is embedded through ViT and BERT following MLP. Afterward, the context grounding module aggregates all features and is used to predict the next step label.}
  \label{fig1}
\end{figure*}
In this section, we present our approach to deal with Affordance-centric Question-driven Task Completion(AQTC) task in detail. We have three major components as shown in Figure \ref{fig1} and each component will be described below.\\
\subsection{Feature Extraction}
We adopt the baseline's strategy\cite{wong2022assistq} to encode multiple source data. For the video, we use a vision transformer(ViT)\cite{dosovitskiy2020image} to encode one frame per second, resulting in feature $V$ of shape $f \times d_v$, where $f$ is the number of frame and $d_v$ is video embedding dimension. For the script, we use BERT\cite{devlin2018bert} to embed $e$ sentences and each sentence's pooled output is used as feature $S$ of shape $e \times d_t$, where $d_t$ is text embedding dimension. Question and answer text are also encoded as the script, resulting in feature $Q$ of shape $1 \times d_t$ and $A$ of shape $j_{max} \times d_t$, where $j_{max}$ is the number of buttons. To encode the user image, masked and reverse-masked button images are used to represent button information and fed into ViT to get feature $I$ of shape $j_{max} \times d_v$. Every feature passes the MLP before the context grounding module.
\subsection{Context Grounding Module}
We modulate the context grounding module to better represent the input features. First, feature $A$ and $I$ are fed into cross attention as they are the closest features to the answer. Then the output is used to calculate cross attention to the feature $Q$ and $S$ respectively to get the feature $F_t$,
\begin{align}\label{eq1}
    F_t = Attn(Attn(Attn(I, A, A), Q, Q), S, S),
\end{align}
where $Attn(query, key, value)$ is the attention operation \cite{bahdanau2014neural}.

\par For the video feature, cross attention with the feature $S$ is calculated as it is the most correlated feature, and multiplied with $W(F_t)$, where $W(F_t)$ refers to the average of multihead attention weight, with shape $f \times e$, 
\begin{align}\label{eq2}
    F_v = W(F_t)^\top Attn(S, V, V). 
\end{align}
\par Finally, as the input feature to the step network, the output of this module is generated as equation (\ref{eq3}). We have explored using other features for the concatenation and could yield better performance by including the feature $I$ and $A$ and we assume it is because they are close features to the answer, as mentioned above.
\begin{align}\label{eq3}
    output = MLP([F_v;F_t;I;A]).
\end{align}
\subsection{Step Network}
Following the baseline, we also explore using both MLP and GRU\cite{chung2014empirical} and utilize previous step's state for the next step prediction. A comparison with different step networks is given in Sec. \ref{ablation}.

\section{Experiments}
\subsection{Data}\label{data}
\begin{figure}[t]
\centering
\includegraphics[height=0.6\linewidth]{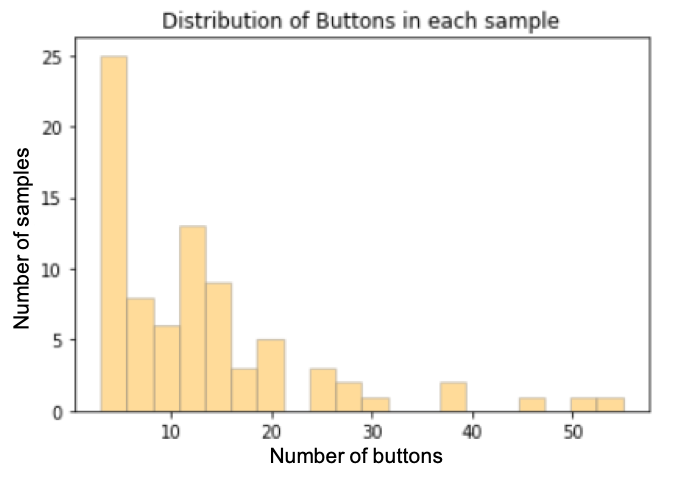}
\caption{Distribution of the number of the buttons}
\label{fig2}
\end{figure}

AssistQ dataset includes a total 100 sets of the instructional videos, scripts, button bounding boxes of each sample. As shown in Figure \ref{fig2}, the distribution of the number of buttons varies. Samples that have less than 10 buttons are almost half of the training set while between 11 to 20 buttons and over 20 buttons have 29 and 12 samples, respectively. 
\par When we split the 80 pairs of training samples into 60:20 for training and validation set to tune model and hyperparameters, we make the validation set to have the balanced number of buttons. Later, all training samples are used to train the model. Test was performed on 20 pairs of instructional egocentric videos without ground truth.

\subsection{Implementation Details}
All of our experiments are conducted in one NVIDIA GeForce RTX 3090. 
In the training stage, we train our model for 100 epochs and use the best among them, with batch size 16, and use Cross-Entropy loss and SGD optimizer with the initial learning rate of 0.002. Both ViT and BERT embed the input data with dimension 768 and the attention operation has 3 heads. Every MLP used for our model have 2-layer mlp with PReLU activation function, following the single neural layer. 

\subsection{Results}
Our best model for the challenge was achieved by using GRU as a step network and by tuning the hyperparameters. As mentioned in Sec. \ref{data}, we utilized a few samples in training set to tune the model as validation set and then make use of all samples when to report for the challenge. The same four evaluation metrics as Visual Dialog \cite{alamri2019audio, das2017visual} are used and the performance for each metric is given in Table \ref{table1} along with the results from the baseline model \cite{wong2022assistq}. All metrics except MR should pursue higher. The best model's checkpoint can be found in our Github repository.
\begin{table}[h]
\centering
\caption{Best model's performance}
\begin{tabular}{@{}lllll@{}}
\toprule
     & R@1    & R@3    & MR     & MRR    \\ \midrule
Our best model & \textbf{.38}& \textbf{.75} & \textbf{2.69} & 3.11 \\ \midrule
Baseline \cite{wong2022assistq} & .30 & .62 & 3.22 & \textbf{3.19} \\ \midrule
\end{tabular}
\label{table1}
\end{table}

\subsection{Analysis and ablation study}\label{ablation}
With the observation of the number of buttons in Sec. \ref{data}, we investigate the impact of the number of buttons on the model's performance. For the experiment, we divide the test set into three subsets as Table \ref{table3} and evaluate our best model on each subset. 
\par As a result, the model could achieve higher score on samples with fewer number of buttons, surprisingly high for the Recall@3 and MR. Meanwhile, performance degrades drastically as the number of buttons gets bigger, except for MRR, suggesting that the model would gain performance by having ability to handle the samples with lots of buttons.
\begin{table}[]
\centering
\caption{Our best model's performance on different test subsets}
\begin{tabular}{@{}lllll@{}}
\toprule
     & R@1    & R@3    & MR     & MRR    \\ \midrule
\# of buttons $\leq$ 10 & \textbf{.41} & \textbf{.91} & \textbf{2.01} & 2.25 \\ \midrule
10 $<$ \# of buttons $\leq$ 20 & .40 & .65 & 2.90 & 3.75 \\ \midrule
20 $<$ \# of buttons & .17 & .50 & 4.17 & \textbf{4.17} \\ \midrule
all samples & .38& .75 & 2.69 & 3.11 \\ \midrule
\end{tabular}
\label{table3}
\end{table}

\par We also conducted an ablation study with different step networks and video features. First we replace GRU with MLP for the step networks and could see that the GRU performs better, suggesting that the recurrent units are helpful for AQTC task. Sampling rate was doubled to augment the video feature but the original tends to be better, and we assume that as the AssistQ dataset consists of the relatively static videos, higher sampling rate does not guarantee the richer information. 


\section{Conclusion}
In our model, we propose a new context grounding module for AQTC task and it shows that a better representation of data could yield better performance. We also utilized a portion of training set to make a validation set, to accurately tune the model and hyperparameters. An analysis over the number of buttons shows that the task get much harder as the number of buttons increases. An ablation study on step network and video feature is conducted and shows the effectiveness of different step networks and video sampling rates. As a result, we won the 2nd place in CVPR'22 LOVEU competition with 1st place in two out of four evaluation metrics.
\section{Acknowledgement}
This work was supported by Pyler, Inc. (video representation learning for context understanding) and Institute of Information \& communications Technology Planning \& Evaluation(IITP) grant funded by the Korea government(MSIT) (No.2020-0-01336, Artificial Intelligence Graduate School Program(UNIST)).

{\small
\bibliographystyle{ieee_fullname}
\bibliography{reference}
}

\end{document}